%% file: main.tex
\pdfoutput=1
\documentclass[11pt]{article}

\usepackage[final]{acl}

\usepackage{times}
\usepackage{latexsym}

\usepackage[T1]{fontenc}
\usepackage[utf8]{inputenc}

\usepackage{microtype}

\usepackage{inconsolata}

\usepackage{graphicx}
\usepackage[utf8]{inputenc} % allow utf-8 input
\usepackage[T1]{fontenc}    % use 8-bit T1 fonts
\usepackage{hyperref}       % hyperlinks
\usepackage{url}            % simple URL typesetting
\usepackage{booktabs}       % professional-quality tables
\usepackage{amsfonts}       % blackboard math symbols
\usepackage{nicefrac}       % compact symbols for 1/2, etc.
\usepackage{xcolor}         % colors
\usepackage{enumitem}       % 没有缩进的enumrate和item
\usepackage{subcaption}     % subfigure
\usepackage{float}          % for [H]
\usepackage{multirow}       % 用于合并行
\usepackage{colortbl}
\usepackage{amsmath}
\usepackage{amssymb}
\usepackage{amsthm}
\usepackage{bbm}            % Indicate (1)
\usepackage{algorithm}      % psudocode
\usepackage{algpseudocode}
\usepackage[normalem]{ulem}
\usepackage[marginal]{footmisc}
\usepackage{algorithm}
\usepackage{algpseudocode}

\usepackage[capitalise]{cleveref}

\usepackage{xspace}
\makeatletter
\DeclareRobustCommand\onedot{\futurelet\@let@token\@onedot}
\def\@onedot{\ifx\@let@token.\else.\null\fi\xspace}

\def\eg{\emph{e.g}\onedot} 
\def\ie{\emph{i.e}\onedot}

\def\wrt{w.r.t\onedot}

\makeatother

\DeclareMathOperator*{\argmax}{arg\,max}

\title{Faster and Better LLMs via Latency-Aware Test-Time Scaling}

\author{Zili Wang$^{1,2}$\thanks{Equal contribution.}, 
  Tianyu Zhang$^{3*}$,  
  Haoli Bai$^{3}$,
  Lu Hou$^{3}$, 
  \\
  \bf Xianzhi Yu$^{3}$, 
  Wulong Liu$^{3}$,
  Shiming Xiang$^{1,2,\dagger}$,
  Lei Zhu$^{3,\dagger}$
  \\
  $^1$ School of Artificial Intelligence, University of Chinese Academy of Sciences, China\\
  $^2$ MAIS, Institute of Automation, Chinese Academy of Sciences, China\\
  $^3$ Huawei Noah’s Ark Lab\\
  wangzili2022@ia.ac.cn,\quad \{zhangtianyu59,zhulei168\}@huawei.com
  }

\begin{document}
% \maketitle

% \input{sec/temp_outlines}
% \newpage

\input{tex_figure/teaser}

\footnote{\quad\quad$^*$ Equal Contributions.}
\footnote{\quad\quad$^\dagger$ Corresponding Author.}
% \vspace{-4em}
\input{sec/0_abstract}

\input{sec/1_introduction}

\input{sec/2_related_work}

\input{sec/3_methodology}

\input{sec/4_experiments}

\input{sec/5_conclusion}

\bibliography{reference}

\appendix

\label{sec:appendix}
\input{sec/X_appendix}

\end{document}

%% file: tex_figure/teaser.tex
\twocolumn[{%
    \renewcommand\twocolumn[1][]{#1}%
    % \vspace{-2em}
    \maketitle
    \vspace{-1cm}
    \begin{center}
        \centering
        \includegraphics[width=0.85\textwidth]{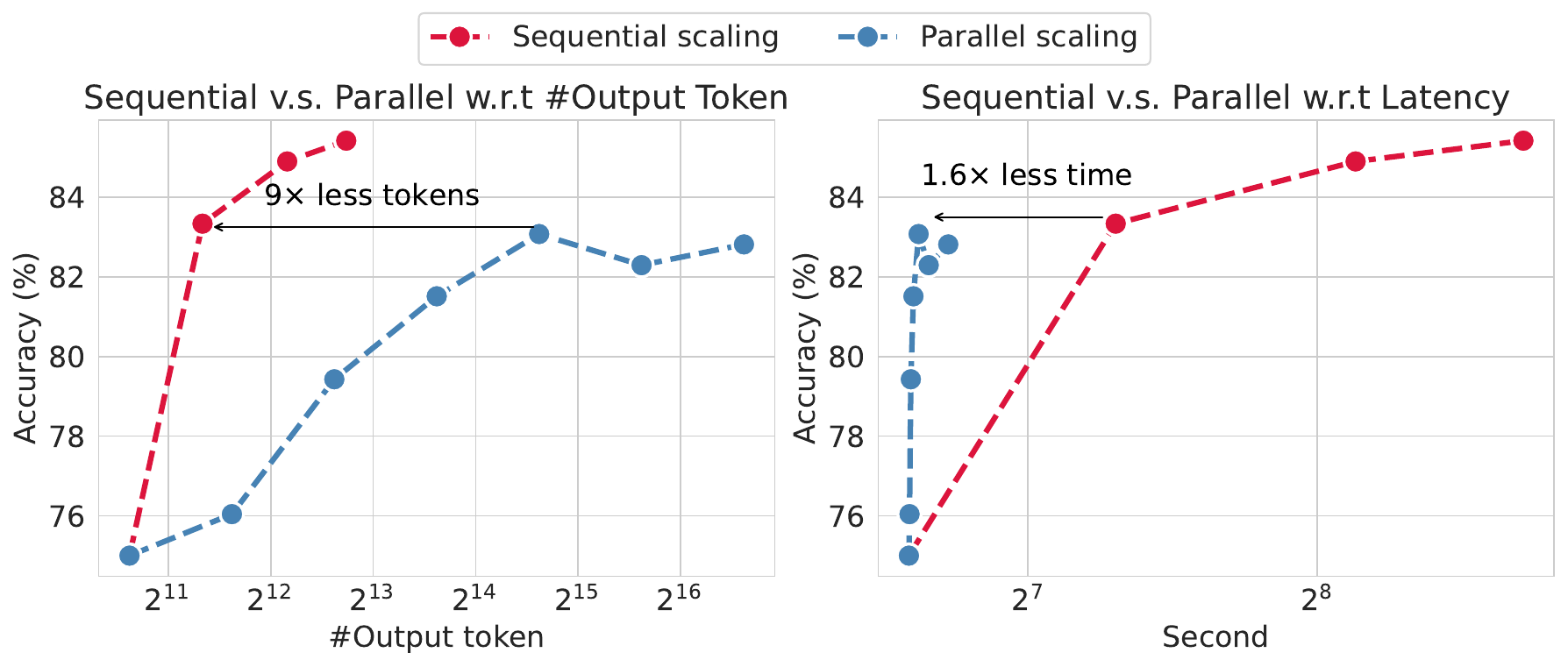}
        % \vspace{-0.2cm}
        \captionof{figure}{\small ``Compute-optimal'' does not necessarily translate to ``latency-optimal'' for test-time scaling.
\textbf{Left:} Previous works measure test-time scaling by \#token budget, indicating sequential scaling shows superior token efficiency than parallel scaling (majority voting for instance). 
\textbf{Right:} When considering \textit{latency} as budget, parallel scaling can be 1.6x faster to achieve the same accuracy than sequential scaling.
% with $1.6\times$ less time. 
Experiments are performed using s1.1-32B~\cite{muennighoff2025s1} on MATH-500~\cite{hendrycks2021math500}.
The red curve corresponds to sequential scaling with sequence length varying in \{1024, 2048, 4096, 8192\}, via budget forcing.
The blue curve represents parallel scaling by majority voting with a fixed sequence length.
% The x-axis uses a logarithmic scale.
        }
        \label{fig:teaser}
        \vspace{0cm}
    \end{center}
    }]

%% file: sec/0_abstract.tex
\begin{abstract}

Test-Time Scaling (TTS) has proven effective in improving the performance of Large Language Models (LLMs) during inference. 
However, existing research has overlooked the efficiency of TTS from a latency-sensitive perspective. Through a latency-aware evaluation of representative TTS methods, 
we demonstrate that a compute-optimal TTS does not always result in the lowest latency in scenarios where latency is critical. 
To address this gap and achieve latency-optimal TTS, we propose two key approaches by optimizing the concurrency configurations: (1) branch-wise parallelism, which leverages multiple concurrent inference branches, and (2) sequence-wise parallelism, enabled by speculative decoding.
By integrating these two approaches and allocating computational resources properly to each, our latency-optimal TTS enables a 32B model to reach 82.3\% accuracy on MATH-500 within 1 minute and a smaller 3B model to achieve 72.4\% within 10 seconds. 
Our work emphasizes the importance of latency-aware TTS and demonstrates its ability to deliver both speed and accuracy in latency-sensitive scenarios.

% 
% This paper investigates two primary research questions: 1. How to improve latency-aware TTS? 2. How to improve latency-aware TTS, and can they enable models to solve problems within a short latency (\eg, 1 minute)?
% % 
% Through comprehensive experiments, we discover that the optimal TTS strategy largely depends on the inherent TTS characteristics of the LLM itself. Furthermore, our proposed latency-optimal strategies enable a 32B model to solve problems within 1 minute, and a smaller 3B model to do so within 10 seconds. Our findings emphasize the necessity of re-evaluating TTS efficiency in latency-sensitive scenarios and demonstrate the feasibility of optimizing it to operate 
% % within the human-acceptable time period.
% to satisfy latency constraints.
\end{abstract}

%% file: sec/1_introduction.tex
\input{tex_figure/different_model}
\section{Introduction}

Test-Time Scaling (TTS) is an effective approach to improve the performance of Large Language Models (LLMs) at the cost of additional inference-time computations~\cite{snell2024scaling,brown2024largelanguagemonkey}.
TTS can be realized by two basic approaches: sequential scaling and parallel scaling. 
Sequential scaling requires the model to produce an extended reasoning process in a single pass~\cite{muennighoff2025s1}. In contrast, parallel scaling generates multiple solutions in parallel and selects the final answer, usually through majority voting~\cite{wang2022majorityvote,liu2025can1b}. Hybrid approaches can be constructed on top of the two basic ones~\cite{guan2025rstar,wang2024moa}.

With the number of generated tokens (\#tokens) as budget, many existing studies~\cite{snell2024scaling,setlur2025rlsub,yang2025towardsthink,liu2025can1b,shi2025heimdall,zhang2024generativeverifier} have examined compute-optimal strategies that enhance average performance gain per token, a metric we refer to as \emph{token efficiency}.
However, in latency-sensitive scenarios where small batch sizes are employed, \eg, personal computer, small-scale commercial deployment and edge device, \emph{``compute-optimal'' does not necessarily translate to ``latency-optimal''}.
This discrepancy is because the performance achieved within a limited time is determined by token efficiency as well as the throughput (\ie, average number of output tokens per second).
As shown in Figure~\ref{fig:teaser}, for s1.1-32B~\cite{muennighoff2025s1} model, although sequential scaling has a better token efficiency, achieving a similar performance with 9$\times$ fewer tokens compared to parallel scaling, it turns out that parallel scaling achieves 1.6$\times$ lower latency.
In fact, under small batch sizes, the time of an LLM autoregressive decoding step is dominated by the memory access of the parameters. Therefore, a moderately increased number of parallel branches incurs little additional latency, allowing a much higher throughput almost for free, as shown in Figure~\ref{fig:roofline}.

\input{tex_figure/roofline}
We then ask \emph{how we can achieve latency-optimal test-time scaling}. 
One lesson from the observation above is that, in addition to optimizing token efficiency, attention must be given to improving the generation concurrency to chase a throughput.
There are two approaches to improving concurrency: (1) branch-wise parallelism, which increases the number of parallel branches $B$, and (2) sequence-wise parallelism, which generates multiple successive tokens for a sequence in a single forward pass with speculative decoding~\cite{leviathan2023fast,chen2023accelerating}.
However, current research lacks a thorough analysis of how to allocate computational power to these two resource-competing approaches and to what extent they can be improved.
% 
% This limitation makes it difficult for users to select TTS models with superior user experience and faster response, and a comprehensive TTS analysis with system-level attributes needs to be conducted.
% 
% This gap makes it challenging for users to \raym{configure inference strategies} that optimally balance response speed and quality, underscoring the need for comprehensive system-level evaluation of test-time scaling attributes.

To bridge the gap, we examine the impact of concurrency configuration for latency-aware test-time scaling in this study.
% 
% Furthermore, we \zilim{systematically evaluate performance} limits and show the huge potential of \emph{latency-optimal TTS}. 
% 
% \raym{Furthermore, we also investigate the potential performance within a set latency and explore strategies to achieve it.}
% 
Specifically, we conduct experiments on representative datasets including MATH-500~\cite{hendrycks2021math500}, AIME24~\cite{aime24}, AIME25~\cite{aime25}, and GPQA-Diamond~\cite{rein2024gpqa} across model sizes from 3B to 32B 
% under varied throughput configurations.
% 
and under varied concurrency configurations.
Revealed by our experiments and analyses, the latency-optimal concurrency depends on the comparative advantage of token efficiency in sequential scaling and parallel scaling.
For models with higher token efficiency with parallel scaling (\eg, LLaMa-3.1-8B-Instruct~\cite{llama3}), one should prioritize branch-wise parallelism.
Otherwise (\eg, for QwQ-32B~\cite{qwq32b}) sequence-wise parallelism takes priority.
To determine the latency-optimal configuration, we propose a simple yet effective greedy search algorithm with less searching steps.
Under 1 minute latency constraint, an optimal concurrency configuration can provide up to 7.3\% better performance than baselines where only a single parallelism is applied.

Our contributions can be summarized as follows:
\begin{enumerate}
%[topsep=3.5pt,itemsep=3pt,leftmargin=20pt]
% \item We introduce the necessity of \zilim{not only improving the token efficiency of certain method, but also \raym{considering latency in test-time scaling}, which is overlooked by previous works.}
\item We introduce latency-aware test-time scaling, which considers the performance scaling under latency constraints. Unlike existing works that prominently only focus on token efficiency, we discover the necessity of taking systemic throughput into consideration for a better latency-performance trade-off.

\item We provide a unified view for parallel branches and speculative decoding from the perspective of generation concurrency. This allows us to frame latency-optimal test-time scaling as a resource allocation problem. A greedy search algorithm is proposed to search the latency-optimal configuration.

\item Through extensive experiments, we explore the optimal concurrency configuration for latency-aware test-time scaling. Our experiments reveal that for s1.1-32B, an optimal concurrency configuration can improve the accuracy by 7.3\% while reducing latency by 1.7$\times$, reaching 82.3\% on MATH-500 in 1 minute.

\end{enumerate}

%% file: tex_figure/different_model.tex
\begin{figure*}[t] % [H] 强制放在这里, [htbp] 是更常用的选项 (here, top, bottom, page)
\centering
\includegraphics[width=0.95\linewidth]{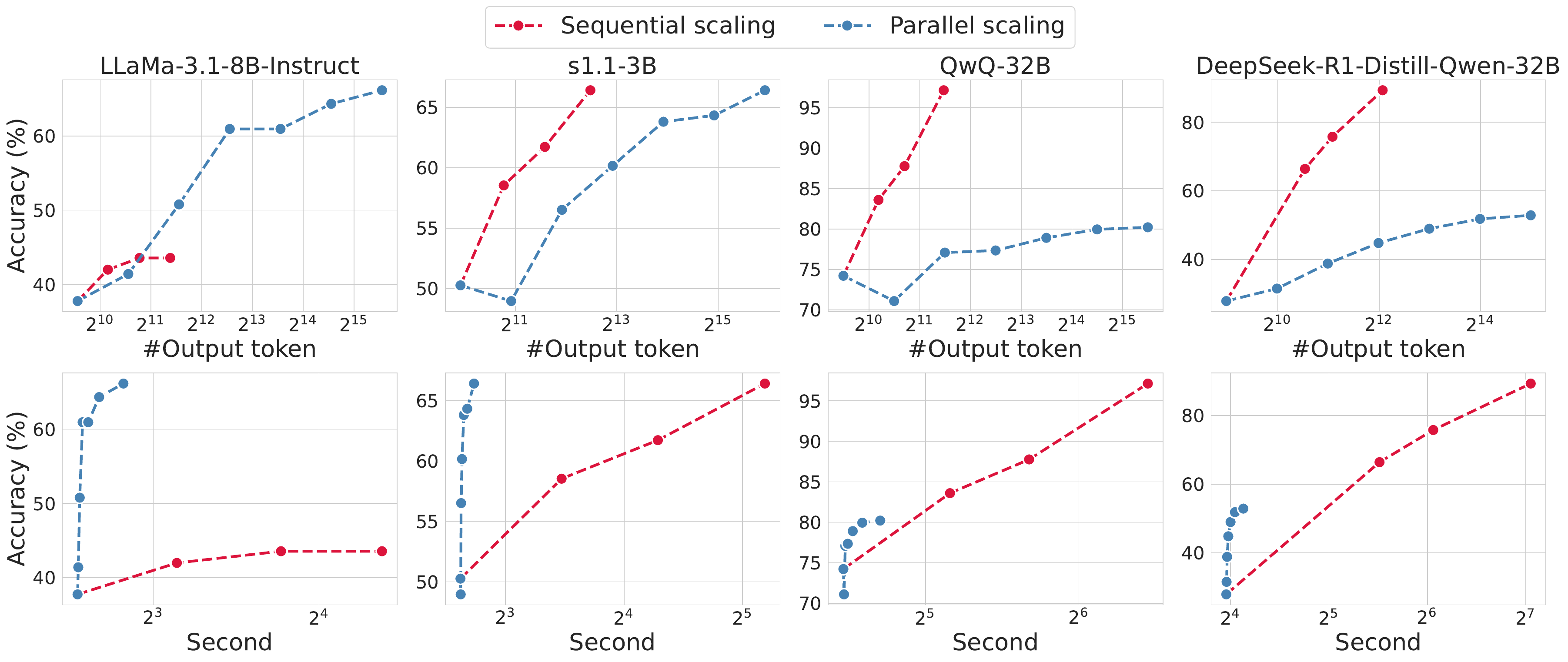}
\caption{Latency-aware test-time scaling on MATH-500 with different model types, with sequential scaling in red and parallel scaling in blue.} % 整个 Figure 的总标题
\label{fig:different} % 整个 Figure 的标签

\end{figure*}

%% file: tex_figure/roofline.tex
\begin{figure*}[t] % [H] 强制放在这里, [htbp] 是更常用的选项 (here, top, bottom, page)
\centering
\includegraphics[width=\linewidth]{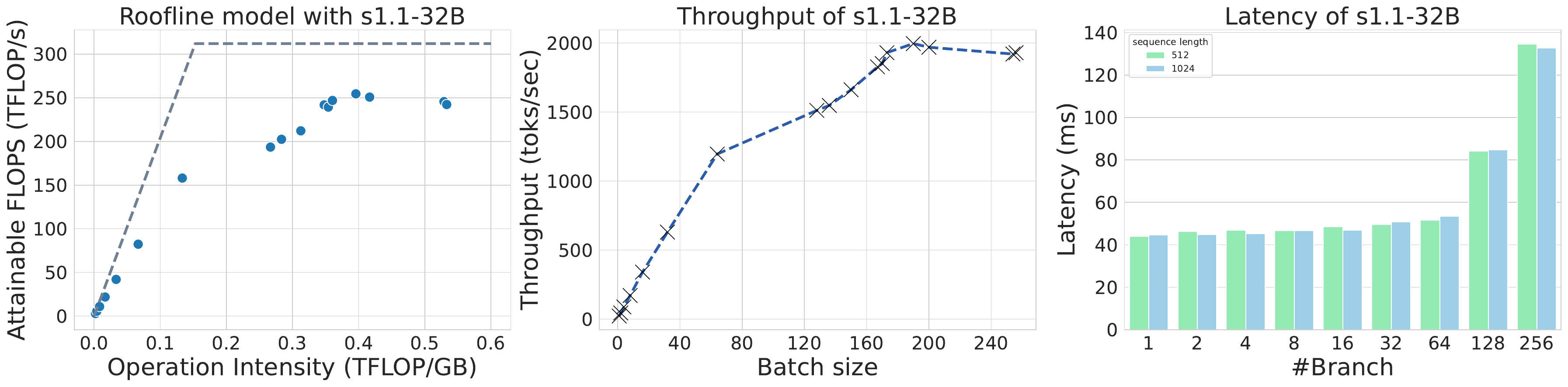}
\caption{ \small System state, latency, and throughput of s1.1-32B. \textbf{Left}: The roofline model with s1.1-32B. Increasing computational demand shifts execution from memory-bound to compute-bound.
\textbf{Middle}: Throughput scales linearly with batch size before saturating at peak FLOPS.
\textbf{Right}: The latency per forward pass under varying batch sizes.
} % 整个 Figure 的总标题
\label{fig:roofline} % 整个 Figure 的标签

\end{figure*}

%% file: sec/2_related_work.tex
\section{Related Work}
\paragraph{Test-time scaling} Scaling the compute at inference time has been proven to be a prominent approach to improve LLM performance. Generally, test-time scaling methods fall into two main categories: \textbf{sequential} and \textbf{parallel} scaling. 
For the former, sequential scaling methods, represented by OpenAI's o1~\cite{o1}, DeepSeek-R1~\cite{guo2025deepseekr1}, and Qwen QwQ~\cite{qwq32b}, enforce the model to generate longer solutions with a detailed reasoning chain.
Such an ability can be incentivized with supervised training~\cite{nye2021Scratchpads,lee2025evolving,muennighoff2025s1} or reinforcement learning~\cite{guo2025deepseekr1}.
Another main category is parallel scaling. This method first generates multiple response candidates in parallel, then applies a selection criterion to identify the best output.
Recent research primarily uses token count as a metric to measure the budget of test-time scaling~\cite{snell2024scaling,setlur2025rlsub,yang2025towardsthink}. Recent work~\cite{singhi2025solve} shows that majority voting outperforms verifier-based methods under low/medium token budgets. Notably, low latency necessitates limited token usage, where majority voting is superior. Thus, our work employs majority voting for parallel scaling. Different from previous works, our work points out that compute-optimal does not necessarily translate to latency-optimal. We introduce latency-aware test-time scaling and figure out the concurrency configuration to latency-optimal TTS. 

\paragraph{Speculative decoding}
Speculative decoding uses a small draft model to generate several draft tokens autoregressively, and the target model to verify them in parallel while ensuring a lossless acceleration~\cite{leviathan2023fast,chen2023accelerating}. Many studies focus on improving acceleration. EAGLE-series~\cite{li2024eagle,li2024eagle2,li2025eagle3}, Medusa~\cite{cai2024medusa}, and Hydra~\cite{ankner2024hydra} employ features from the target model for better draft acceptance rate. HASS~\cite{zhang2025hass} mitigates the inconsistency between training and inference by simulating the multi-step draft generation in the training phase. Hierarchical Drafting~\cite{cho2025HR} improves the acceptance rate by regarding the hierarchical drafting strategy based on temporal locality. MoESD~\cite{huang2025moesd} shows that at medium batch sizes, MoE models with speculative decoding can surpass dense models. Our work demonstrates the feasibility of latency-aware TTS optimization through speculative decoding, presenting it as one viable approach to latency-optimal TTS.

%% file: sec/3_methodology.tex
\section{Rethinking Test-Time Scaling with Latency}

Previous works measure the budget of LLM test-time scaling by \#tokens. However, it provides an incomplete evaluation, particularly when considering real-world deployment constraints. Our investigation reveals the necessity to consider latency-aware TTS in latency-sensitive scenarios.
\input{tex_figure/overview}
\subsection{Latency-Aware Test-time Scaling}

The inference of LLMs on modern accelerators is a memory-bound process, constrained by memory bandwidth. Nowadays, LLMs have billions of parameters~\cite{qwen,qwen2.5,qwq32b,guo2025deepseekr1,llama3}. In the autoregressive decoding, parameters (weights \& KV caches) are loaded from memory (e.g., HBM on a GPU) into the compute units (e.g., SMs on a GPU) for matrix multiplication and other arithmetic operations. The runtime paid for the iteration is dominated by memory access. This is called the memory-bound nature of LLM. Under memory-bound constraints, slightly increasing the computation overhead does not affect the inference latency, but can increase throughput, as shown in Figure~\ref{fig:roofline}. 

To characterize latency-aware test-time scaling under the memory-bound scenario, we explicitly consider two key factors:

\textbf{Token efficiency} is the ratio of task-specific accuracy improvement per token generated by the LLM, measured by \%/token. A high token efficiency indicates that each token contributes significantly to achieving the accuracy. Prior work's approach to determining compute-optimal TTS using \#token as budget fundamentally represents a search for maximal token efficiency.

\textbf{Throughput} is the number of output tokens generated per wall-clock time, measured by token/s. Note that the throughput may vary as the sequence length grows. A high throughput indicates efficient use of hardware resources and the system's capacity to quickly handle a large volume of tokens.

Consider s1.1-32B~\cite{muennighoff2025s1} as an example. Figure~\ref{fig:teaser} compares the accuracy-\#token and accuracy-latency curves of sequential and parallel scaling. Sequential scaling achieves higher token efficiency (83.3\%/$2^{11.3}$toks) than parallel scaling (83.3\%/$2^{14.4}$toks), but Figure~\ref{fig:roofline} reveals its throughput is $16\times$ lower. Consequently, sequential scaling requires $1.6\times$ more time to attain comparable accuracy. This suggests that while sequential scaling improves token efficiency, its inferior throughput hinders better accuracy in limited time. Thus, test-time scaling budgets must consider both token efficiency and throughput to achieve high accuracy with low latency.

\subsection{How to Improve Latency-Aware TTS? A Concurrency Perspective}

To improve latency-aware TTS, the key insight is to improve generation concurrency to increase throughput. To this end, there exist two approaches from the perspective of concurrency: 

\paragraph{Branch-wise Parallelism.}
One approach is employing multiple concurrent branches $B$ for the question, as shown in Figure~\ref{fig:overview} (b). For instance, when a 2048-token response fails to yield a correct answer, users can infer more branches to generate multiple responses of the same length and determine the final answer through majority voting. This approach harnesses more underutilized memory-bound computational resources, introducing almost no extra latency. Employing multiple branches to explore diverse reasoning paths brings further improvements of TTS performance.

\paragraph{Sequence-wise Parallelism}
Another effective approach is speculative decoding~\cite{leviathan2023fast,chen2023accelerating}, as shown in Figure~\ref{fig:overview} (c). SD accelerates by verifying multiple draft tokens concurrently, leveraging underutilized memory-bound computational resources to mitigate the memory access burden in sequential generation with lossless performance. With SD, LLM can generate longer responses within a limited time, thereby enhancing TTS performance.

The combination of the two approaches is illustrated in Figure~\ref{fig:overview} (d). Branch-wise parallelism enhances performance without increasing latency, elevating the scaling curve. Sequence-wise parallelism reduces latency without compromising performance, causing the scaling curve to shift leftward. Their combined effect moves the scaling curve toward the upper-left quadrant. 

\subsection{Latency-Optimal Test-Time Scaling}

The joint application of both approaches increases the overall concurrency and introduces additional computational overhead. When this overhead surpasses memory access overhead, the system transitions into a compute-bound state, leading to more latency. Besides, branch-wise parallelism exhibits diminishing improvement with increasing branches, while sequence-wise parallelism's acceleration reaches a maximum threshold. Consequently, an optimal boundary curve represents the latency-optimal test-time scaling, as shown by the green curve in Figure~\ref{fig:overview} (d). This curve defines where neither latency nor accuracy can be further improved without concurrency trade-offs.

We aim to determine the latency-optimal TTS strategy by allocating concurrency resources with parallel branches $B$ and draft length $\gamma$. Let $\text{Target}(\theta,T,x)$ be the output distribution over problem $x$ produced by the LLM with test-time compute hyperparameter $\theta$ and time limitation $T$. The settings of branches and draft length are included by $\{B, \gamma\}\subsetneq\theta$. The latency-optimal test-time scaling is given by:
\begin{align}
&\theta^*_{x,y^*(x)}(T) =\notag\\
&\argmax_\theta \left(\mathbb{E}_{y\sim\text{Target}(\theta,T,x)}\left[\mathbbm{1}_{y=y^*(x)}\right]\right),
\label{eqn:formualte}
\end{align}
where $y^*(x)$ indicates the groundtruth of corresponding problem $x$, and $\theta^*_{x,y^*(x)}(T)$ represents the test-time latency-optimal scaling strategy for problem $x$ with time $T$. Finding the optimal $\theta^*_{x,y^*(x)}(T)$ is also the way to find optimal branches $B^*$ and draft length $\gamma^*$. 

\input{tex_table/greedy_search_alg}

Finding $(B^*,\gamma^*)$ requires to search the configuration space. Although grid search can find the optimal configuration, it brings a significant cost to search the entire configuration space. Therefore, we introduce a simple yet effective greedy search algorithm, as shown by Algorithm~\ref{alg:greedy}.

In practice, $B$ takes values in powers of 2 ($B = 2^k$), while $\gamma$ increments in steps of 1. The basic grid search evaluates every $(B,\gamma)$, with complexity of $O((\log B_{max}+1)\times \gamma_{max})$. With greedy search, the complexity reduces to  $O((\log B^*+1)+\gamma^*)$, where $B^*$ and $\gamma^*$ denote the optimal configuration.

\input{tex_figure/branch_tradeoff}
\input{tex_figure/sd_tradeoff}

\input{tex_figure/latency_optimal_1}
\input{tex_figure/latency_optimal_2}

%% file: tex_figure/overview.tex
\begin{figure*}[h] % [H] 强制放在这里, [htbp] 是更常用的选项 (here, top, bottom, page)
\centering
\includegraphics[width=0.9\linewidth]{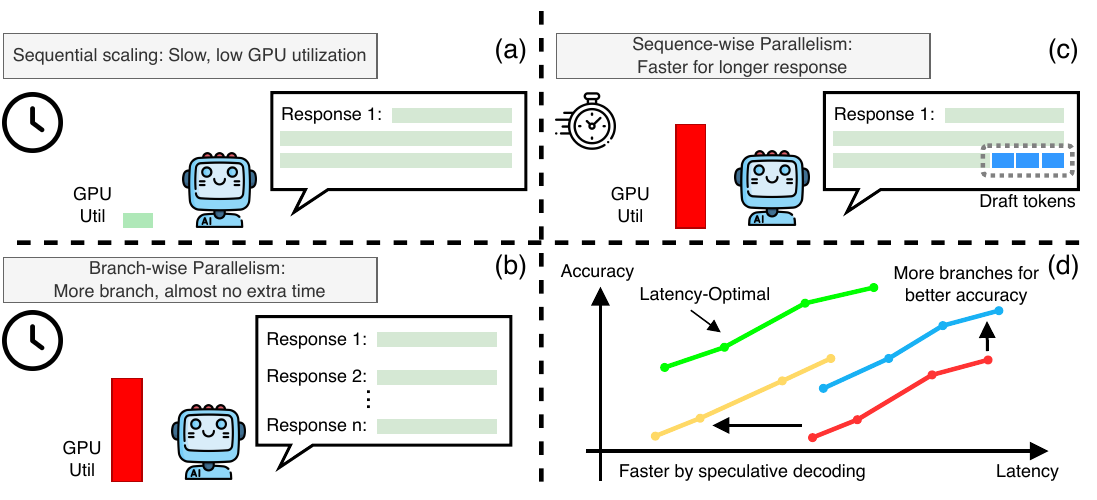}
\caption{Overview of how to improve TTS with latency budget. (a): default sequential scaling suffers from long latency due to low GPU utilization. (b): Branch-wise parallelism employs more branches to utilize FLOPs, improving accuracy with no extra time. (c): Sequence-wise parallelism utilizes FLOPs with speculative decoding, reaching faster generation and theoretically no performance loss. (d): With speculative decoding, the curve (\textcolor[rgb]{1,0.2,0.2}{red}) shifts to the left (\textcolor[rgb]{1,0.85,0.4}{yellow}), indicating reduced latency. With more branches, the curve shifts upward (\textcolor[rgb]{0.1,0.7,0.956}{blue}), indicating improved performance. Latency-optimal TTS can be achieved by jointly applying these two optimizations (\textcolor[rgb]{0,1,0}{green}).} % 整个 Figure 的总标题
\label{fig:overview} % 整个 Figure 的标签

\end{figure*}

%% file: tex_table/greedy_search_alg.tex
\begin{algorithm}[h]
\caption{Greedy Search for Latency-Optimal TTS}
\begin{algorithmic}[1]
\footnotesize % <-- 在这里添加 \footnotesize 命令
\Require $tts\_task$: TTS evaluation. $T$: latency budget. $(B, \gamma)$: configuration; $acc$: accuracy

\State $B \gets 1$
\State $\gamma \gets 0$
\State $acc \gets tts\_task(B, \gamma, T)$ \Comment{Baseline accuracy}
\State $end \gets \textbf{False}$

\While{not $end$}
    \State $acc_b \gets tts\_task(2B, \gamma, T)$ \Comment{Expand branch-wise}
    \State $acc_\gamma \gets tts\_task(B, \gamma + 1, T)$ \Comment{Expand sequence-wise}
    
    \If{$acc_b \geq acc_\gamma$}
        \State $B_{new}, \gamma_{new}, acc_{new} \gets 2B,\gamma, acc_b$
    \Else
        \State $B_{new}, \gamma_{new}, acc_{new} \gets B,\gamma+1, acc_\gamma$
    \EndIf

    \If{$acc_{new} > acc$}
        \State $B, \gamma, acc \gets B_{new},\gamma_{new}, acc_{new}$
    \Else
        \State $end \gets \textbf{True}$
    \EndIf
\EndWhile

\State \Return $(B, \gamma), acc$
\end{algorithmic}\label{alg:greedy}
\end{algorithm}

%% file: tex_figure/branch_tradeoff.tex
\begin{figure*}[t] % [H] 强制放在这里, [htbp] 是更常用的选项 (here, top, bottom, page)
\centering
\includegraphics[width=0.9\linewidth]{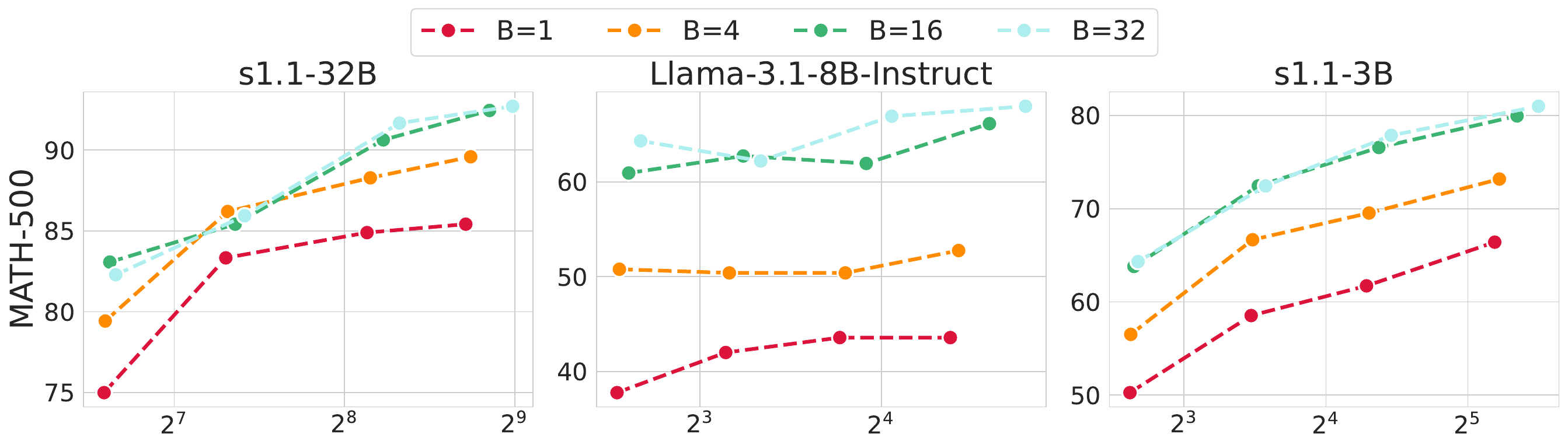}
\caption{Latency-aware test-time scaling with different branches on s1.1-32B, LLama-3.1-8B-Instruct and s1.1-3B. Speculative decoding is not implemented. Each picture shows the scaling curve with varying branch numbers.} % 整个 Figure 的总标题
\label{fig:branch} % 整个 Figure 的标签

\end{figure*}

%% file: tex_figure/sd_tradeoff.tex
\begin{figure}[t] % [H] 强制放在这里, [htbp] 是更常用的选项 (here, top, bottom, page)
\centering
\includegraphics[width=0.9\linewidth]{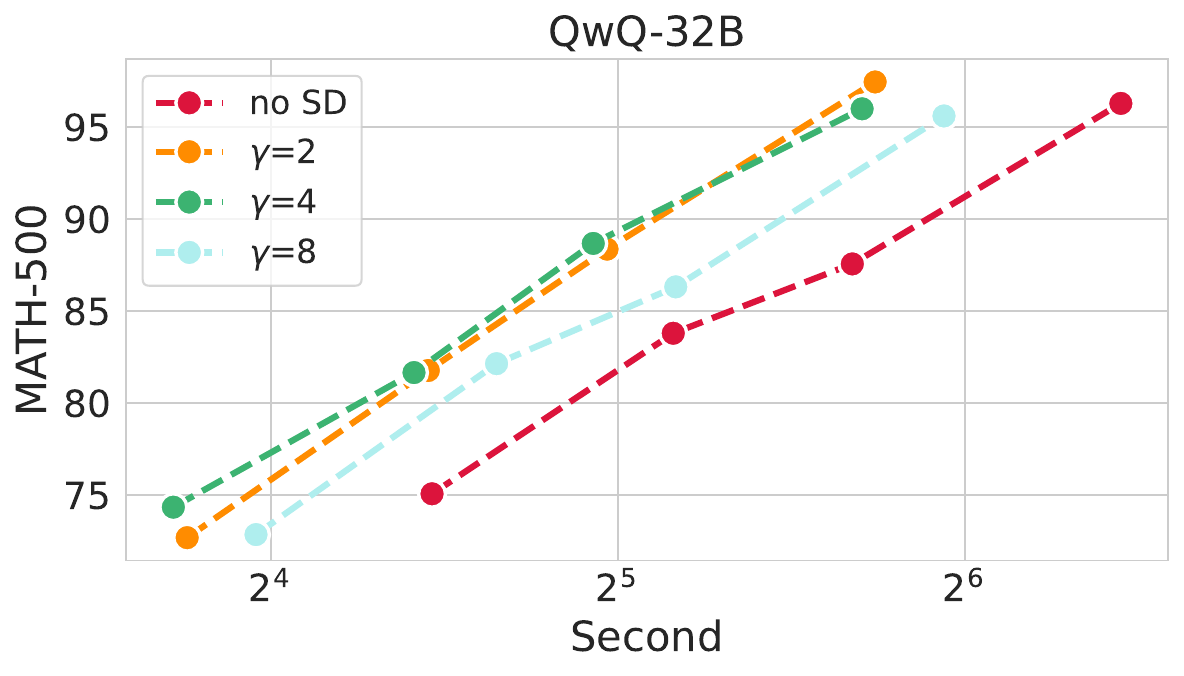}
\caption{Latency-aware test-time scaling with speculative decoding of different draft length $\gamma$ on MATH-500. Number of branch is fixed to 1. Draft model: DeepSeek-R1-Distill-Qwen-7B. Target model: QwQ-32B.} % 整个 Figure 的总标题
\label{fig:sd} % 整个 Figure 的标签

\end{figure}

%% file: tex_figure/latency_optimal_1.tex
\begin{figure*}[h] % [H] 强制放在这里, [htbp] 是更常用的选项 (here, top, bottom, page)
\centering
\includegraphics[width=0.9\linewidth]{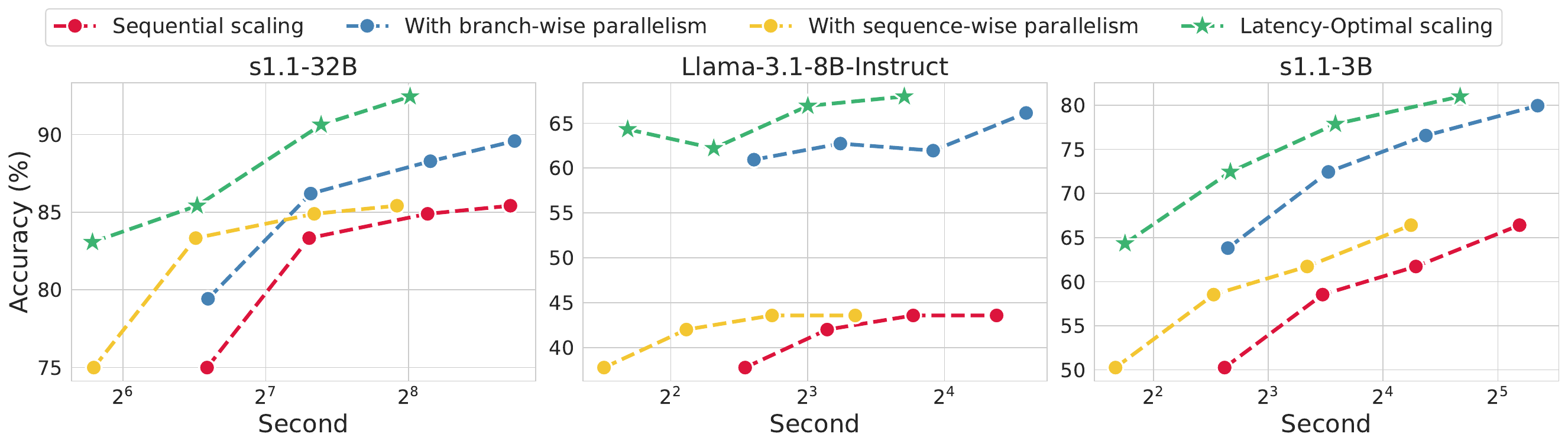}
\caption{Latency-aware test-time scaling curves on MATH-500 with sequential scaling in red, parallel scaling in blue, and latency-optimal in green (the same to subsequent figures).} % 整个 Figure 的总标题
\label{fig:latencyoptimal1} % 整个 Figure 的标签

\end{figure*}

%% file: tex_figure/latency_optimal_2.tex
\begin{figure*}[h] % [H] 强制放在这里, [htbp] 是更常用的选项 (here, top, bottom, page)
\centering
\includegraphics[width=\linewidth]{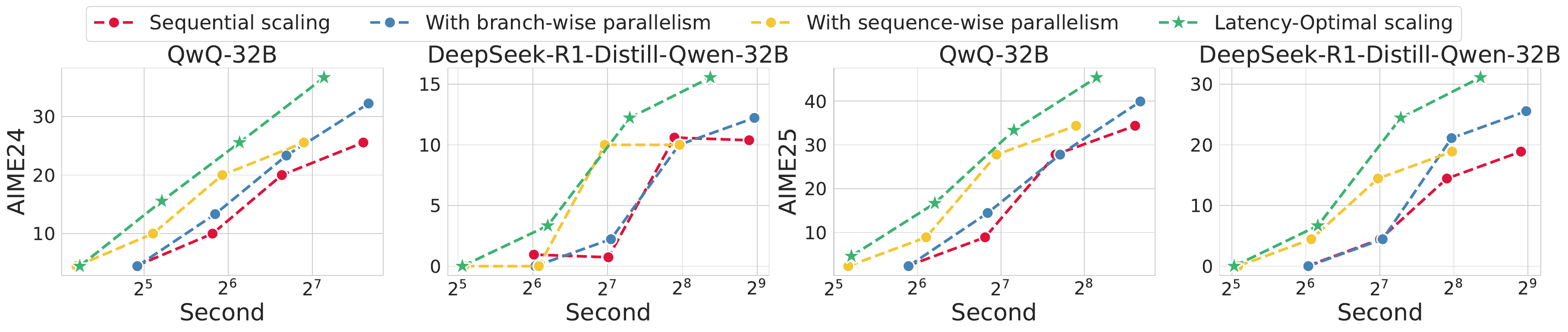}
\caption{Latency-aware test-time scaling curves of QwQ-32B and DeepSeek-R1-Distill-Qwen-32B under latency on AIME24 (left two panels) and AIME25 (right two panels).} % 整个 Figure 的总标题
\label{fig:latencyoptimal2} % 整个 Figure 的标签

\end{figure*}

%% file: sec/4_experiments.tex
\section{Experiments and Discussion}
\subsection{Experimental Setup}

\paragraph{Tasks.} Representative challenging tasks as benchmarks to measure the scaling property are selected: MATH-500~\cite{hendrycks2021math500} is a popular math benchmark comprising 500 high-school competition problems. AIME24~\cite{aime24} and AIME25~\cite{aime25} each consist of 30 math problems from the 2024 and 2025 American Invitational Mathematics Examination (AIME). GPQA-Diamond~\cite{rein2024gpqa} consists of 198 science QA problems encompassing PhD-level physics, chemistry, and biology.

\paragraph{Models.} 
We choose models with parameter sizes suitable for device-side deployment. Also, to implement speculative decoding, we select model types with draft models available. Therefore, considering its scalability, we employ s1.1-32B as our test-time scaling baseline. s1.1-7B is used as its draft model for speculative decoding. Also, we conduct relevant experiments on LLaMa-3.1-8B-Instruct~\cite{llama3} with Eagle3~\cite{li2025eagle3} as the draft model. For RL-based thinking model, we employ DeepSeek-R1-Distill-Qwen-32B~\cite{guo2025deepseekr1} and QwQ-32B~\cite{qwq32b} for their superior reasoning ability. The DeepSeek-R1-Distill-Qwen-7B is used as their draft model. We employ s1.1-3B for small-sized LLM for its outstanding model size and excellent performance with Qwen2.5-0.5B-Instruct~\cite{qwen} as the draft model. To better control the output sequence length, we employ budget forcing~\cite{muennighoff2025s1} as an appropriate method to enforce the model to generate longer CoT. We use majority voting~\cite{wang2022majorityvote} to select the answer. Experiments are conducted on the codebase of OpenR~\cite{wang2024openr}.

\subsection{Sequential and Parallel Scaling under Latency-Aware TTS}

\paragraph{Sequential scaling suffer from low throughput.}
As shown in Figure~\ref{fig:different}, sequential scaling achieves superior performance across most model types. However, its impact is less pronounced for models not explicitly trained for long CoT reasoning, such as LLaMa-3.1-8B-Instruct~\cite{llama3}. Conversely, Figure~\ref{fig:roofline} reveals that sequential scaling exhibits lower throughput. For instance, the s1.1-32B model processes only 22.7 tokens/s, translating to approximately 1,000 tokens per minute. While sequential scaling demonstrates better token efficiency, its lower throughput makes it less effective than parallel scaling under time constraints. Consequently, sequential scaling struggles to generate sufficient tokens to attain high accuracy.

\paragraph{Parallel scaling improves fast, but limited.} In Figure~\ref{fig:different}, parallel scaling shows lower token efficiency than sequential scaling for most models except LLaMa-3.1-8B-Instruct. However, from Figure~\ref{fig:roofline}, increasing the branches of LLM inference hardly requires extra latency. Specifically, when increasing branches from 1 to 16, s1.1-32B gains nearly $10\%$ improvement on MATH-500 at almost no extra latency cost. This can be attributed to the fact that the computation resources are fully utilized. By inference in branches, one autoregressive decoding procedure would generate the branches' tokens, but the latency cost of developing one token by sequential scaling is the same. So parallel scaling can reach a larger \#token faster. When the branches are increased to 64, the performance gain becomes slight, and the latency slightly grows because of the large amount of KV cache. Overall, parallel scaling maximizes hardware parallelism and scales output tokens simultaneously. But, parallel scaling often yields suboptimal accuracy due to branch redundancy and limited scalability.

\paragraph{Mostly, parallel scaling can surpass sequential scaling within limited time period.} As shown in Figure~\ref{fig:different}, merely employing parallel scaling can surpass sequential scaling within a shorter, limited time. However, the extent of performance improvement varies depending on the model type. Specifically, for reasoning models like QwQ-32B, parallel scaling reaches $80.1\%$ accuracy on the MATH-500 dataset within $30$ seconds, but sequential scaling requires $1.4\times$ more time to achieve the comparable performance. Sequential scaling shows a slight performance gain for LLaMa-3.1-8B-Instruct, which is not designed for long CoT reasoning. While parallel scaling is still effective on MATH-500 since it explores more diverse solution paths, it shows obvious token efficiency, and sequential scaling cannot surpass it. Overall, parallel scaling can surpass sequential scaling \wrt latency, reaching a comparable accuracy within a relatively short time.

\subsection{The Impact of Branch-wise Parallelism for Latency-Aware TTS}

We conduct experiments on sequential scaling with varying branch sizes. In these configurations, we implement sequential scaling in parallel, with all parallel branches aggregated through majority voting. As shown in Figure~\ref{fig:branch}, s1.1-32B demonstrates an initial upward trend in the TTS curve as the number of branches increases, indicating effective performance improvements. However, when the branch size grows excessively, the performance gains diminish, and latency increases slightly due to the non-negligible overhead of KV cache. In contrast, LLaMa-3.1-8B-Instruct exhibits minimal sequential scaling effects, resulting in a flatter curve compared to s1.1-32B and QwQ. Nevertheless, parallel scaling proves impactful, yielding an accuracy improvement that elevates the curve.

\input{tex_table/baseline}
\input{tex_figure/32B}
\input{tex_figure/3B}

\subsection{The Impact of Sequence-wise Parallelism for Latency-Aware TTS}

Speculative decoding can accelerate model inference to some extent while preserving accuracy. As shown in Figure~\ref{fig:sd}, most scaling curves with speculative decoding are on the left side of the baseline curve. As the speed-up ratio grows, the left-shifted trend becomes more obvious. However, when the draft length $\gamma$ becomes large, the speed-up ratio would decay, slowing down the overall speed. In our experiments, QwQ-32B is the target model, while DeepSeek-R1-Distill-Qwen-7B as the draft model, achieving a maximum speed-up ratio of $1.64\times$. Therefore, the scaling curve moves to the left first and then to the right. Sequential scaling with speculative decoding can achieve $7.5\%$ higher accuracy than the baseline curve for a limited time. These results demonstrate that employing a proper draft length $\gamma$ can push the inference speed to an optimal stage, enabling further sequential scaling within a limited time.

\subsection{Latency-Optimal TTS with Branch-wise and Sequence-wise Parallelism}

We conduct comprehensive experiments with various parallel branches and draft lengths to identify the concurrency configuration to achieve latency-optimal TTS. The results are shown in Figure~\ref{fig:latencyoptimal1} and~\ref{fig:latencyoptimal2}. For LLMs that benefit primarily from sequential scaling, speculative decoding emerges as the dominant factor in the latency-optimal configuration. Conversely, for LLMs hat exhibit improvements from branches, increasing the number of branches yields huge performance gains. This difference stems from the different token efficiency. For reasoning models like QwQ, accuracy improvements are achieved through long CoT, making better SD acceleration more advantageous within limited time. In contrast, models like LLaMA-3.1-8B-Instruct, which do not benefit from long CoT, using more branches to expand the search paths is more efficient. See appendix for detailed configurations.

\input{tex_table/greedy_search}

\section{Results under Latency-Optimal TTS}

\subsection{Can LLM Solve the Problem in 1 minute?}

Achieving as high accuracy as possible within a limited time holds significant practical value. We aim to find out how TTS can improve the accuracy individually on the device side within a relatively short time limitation, like 1 minute. To this end, we comprehensively evaluate s1.1-32B and s1.1-3B \wrt latency and record their inference latency.

From Figures~\ref{fig:32B} and~\ref{fig:3B}, we derive the following observations: (1) Large model like s1.1-32B, achieves relatively high accuracy within just 1 minute, which is unattainable by the baseline. (2) Smaller model (s1.1-3B) attains notable accuracy in merely 10 seconds, showing significant potential. This trend persists across datasets spanning diverse domains, suggesting that popular models can be optimized using the latency-optimal TTS strategy.

\subsection{How Can Latency-Optimal Improve Compared with Baseline?}
Based on previous findings of latency-optimal TTS with different model types, branches, and draft lengths, we summarize the results in Table~\ref{tab:performance}. For s1.1-32B on MATH-500, we find that latency-optimal TTS can achieve $6\%$ accuracy improvement on average than merely using sequence-wise parallelism, and $1.6\times$ on average faster than branch-wise parallelism. Latency-optimal TTS outperforms the baseline in both accuracy and latency to a noticeable degree. However, increased branch and draft length causes LLMs to enter a compute-bound regime, where computational overhead exceeds memory access overhead. This eliminates the benefits of parallelism, leading to increased latency and sometimes even worse than the baseline. This suggests that a latency-optimal TTS strategy requires extremely fine-grained parameter tuning.

\subsection{Effectiveness of Greedy Search}
To validate the effectiveness of our proposed greedy search algorithm, extensive experiments are conducted across various models on MATH-500. The results in Table~\ref{tab:search_comparison} show that our greedy search achieves the same optimal configuration as grid search, while significantly reducing searching steps (8–10 vs 56). These results show the practical efficiency of our greedy search algorithm.

\input{tex_table/confidence_based}
\subsection{Different Aggregation Strategies}
We conduct additional ablation experiments with s1.1-32B on MATH-500 comparing majority voting (self-consistency) with \textbf{confidence-based aggregation strategies}~\cite{liu2025can1b}: first weight the answers by minimum (min) or average (avg) confidence, then select the answer with highest confidence (max) or accumulates the scores of all identical answers and then selects the answer with the highest score (vote). Results are shown in Table~\ref{tab:aggregation}. Note that confidence-based strategies do not affect the latency. The results show that using confidence scores can indeed slightly improve output quality over simple majority voting.

% intro里提一嘴greedy search的作用是具体投机和并行无关的

%% file: tex_table/baseline.tex
% \begin{table*}[htbp]
% \setlength{\tabcolsep}{12pt}
% \centering
% \caption{Comparison of baseline sequential scaling, speculative decoding, more branches, and latency-optimal TTS with different models and sequence lengths on MATH-500 of s1.1-32B.\todp{discuss on KV cache}}
% \begin{tabular}{cccccc}
% \toprule
% \textbf{SeqLen.} & \textbf{Metrics} & \textbf{Baseline} & \textbf{SpecDec.} & \textbf{Branch} & \textbf{Latency-Optimal TTS} \\
% \midrule
% \multirow{2}{*}{1024} & Latency & 96.2s & 55.5s & 96.7s & 57.2s \\
%  & Accuracy & 75.0\% & 75.0\% & 79.4\% & 82.3\% \\
%  \midrule
% \multirow{2}{*}{2048} & Latency & 157.9s & 91.1s & 159.2s & 95.9s \\
%  & Accuracy & 83.3\% & 83.4\% & 86.2\% & 85.9\% \\
%  \midrule
%  \multirow{2}{*}{4096} & Latency & 280.6s & 161.9s & 284.5s & 179.1s \\
%  & Accuracy & 84.9\% & 84.9\% & 88.3\% & 91.7\% \\
%  \midrule
%  \multirow{2}{*}{8192} & Latency & 419.5s & 241.9s & 428.0s & 283.1s \\
%  & Accuracy & 85.4\% & 85.4\% & 89.6\% & 92.7\% \\
% \bottomrule
% \end{tabular}
% \label{tab:performance}
% \end{table*}

\begin{table*}[t]
\setlength{\tabcolsep}{6pt}
\centering
\begin{tabular}{ccccccccc}
\toprule
\multirow{2}{*}{Strategy} & \multicolumn{2}{c}{1024} & \multicolumn{2}{c}{2048} & \multicolumn{2}{c}{4096} & \multicolumn{2}{c}{8192}\\
\cmidrule(lr){2-3} \cmidrule(lr){4-5} \cmidrule(lr){6-7} \cmidrule(lr){8-9}
 & Lat.~(s) & Acc.~(\%) & Lat.~(s) & Acc.~(\%) & Lat.~(s) & Acc.~(\%) & Lat.~(s) & Acc.~(\%)   \\
 \midrule
\rowcolor{gray!20}\textbf{Baseline}  & 96.2{\scalebox{0.6}{$\pm$0.2}} & 75.0{\scalebox{0.6}{$\pm$0.6}} & 157.9{\scalebox{0.6}{$\pm$0.5}} & 83.4{\scalebox{0.6}{$\pm$1.2}} & 280.6{\scalebox{0.6}{$\pm$0.5}} & 84.9{\scalebox{0.6}{$\pm$1.8}} & 419.5{\scalebox{0.6}{$\pm$0.7}} & 85.4{\scalebox{0.6}{$\pm$0.9}}  \\
\textbf{Bnh-wise}  & 96.7{\scalebox{0.6}{$\pm$0.2}} & 79.4{\scalebox{0.6}{$\pm$0.7}} & 159.2{\scalebox{0.6}{$\pm$0.8}} & 86.2{\scalebox{0.6}{$\pm$1.1}} & 284.5{\scalebox{0.6}{$\pm$0.3}} & 88.3{\scalebox{0.6}{$\pm$1.5}} & 428.0{\scalebox{0.6}{$\pm$0.6}} & 89.6{\scalebox{0.6}{$\pm$1.0}}  \\
\textbf{Seq-wise}  & 55.5{\scalebox{0.6}{$\pm$0.4}} & 75.0{\scalebox{0.6}{$\pm$0.3}} & 91.1{\scalebox{0.6}{$\pm$0.3}} & 83.4{\scalebox{0.6}{$\pm$1.9}} & 161.9{\scalebox{0.6}{$\pm$0.3}} & 84.9{\scalebox{0.6}{$\pm$1.4}} & 241.9{\scalebox{0.6}{$\pm$0.8}} & 85.4{\scalebox{0.6}{$\pm$0.1}}  \\
\rowcolor{gray!20}\textbf{Lat-Opt.}  & 57.2{\scalebox{0.6}{$\pm$0.4}} & 82.3{\scalebox{0.6}{$\pm$1.8}} & 95.9{\scalebox{0.6}{$\pm$0.3}} & 85.9{\scalebox{0.6}{$\pm$0.7}} & 179.1{\scalebox{0.6}{$\pm$0.4}} & 91.7{\scalebox{0.6}{$\pm$1.3}} & 283.1{\scalebox{0.6}{$\pm$0.6}} & 92.7{\scalebox{0.6}{$\pm$0.3}}  \\
\midrule
\textbf{Improve}  & \textcolor[rgb]{0,0,0}{1.7$\times$} & \textcolor[rgb]{0,0,0}{7.3+} & \textcolor[rgb]{0,0,0}{1.6$\times$} & \textcolor[rgb]{0,0,0}{2.5+} & \textcolor[rgb]{0,0,0}{1.6$\times$} & \textcolor[rgb]{0,0,0}{6.8+} & \textcolor[rgb]{0,0,0}{1.5$\times$} & \textcolor[rgb]{0,0,0}{7.3+}  \\
\bottomrule
\end{tabular}
\caption{Results of baseline, branch-wise parallelism, sequence-wise parallelism and latency-optimal TTS with different sequence lengths on MATH-500 of s1.1-32B. Lat.: Latency. Acc.: Accuracy. \textbf{Bnh-wise}: Branch-wise parallelism. \textbf{Seq-wise}: Sequence-wise parallelism. \textbf{Lat-Opt.}: Latency-Optimal. \textbf{Improve} is reported between \textbf{Baseline} and \textbf{Lat-Opt.} Results are obtained from 3 repeated experiments with mean and standard deviation reported.}
\label{tab:performance}
\end{table*}

%% file: tex_figure/32B.tex
\begin{figure*}[h] % [H] 强制放在这里, [htbp] 是更常用的选项 (here, top, bottom, page)
\centering
\includegraphics[width=\linewidth]{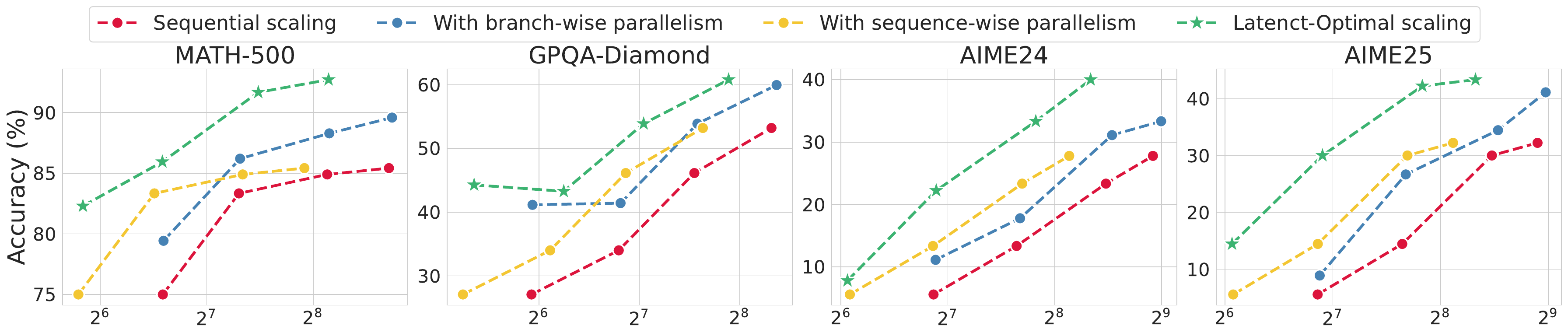}
\caption{Latency-aware TTS curves of s1.1-32B on MATH-500, AIME24, AIME25 and GPQA-Diamond.} % 整个 Figure 的总标题
\label{fig:32B} % 整个 Figure 的标签
\includegraphics[width=\linewidth]{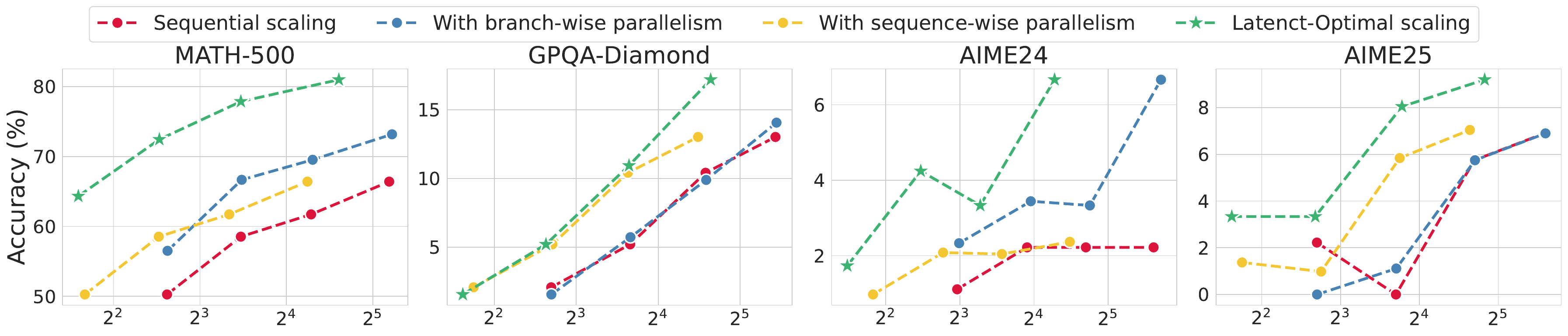}
\caption{Latency-aware TTS curves of s1.1-3B on MATH-500, AIME24, AIME25 and GPQA-Diamond.} % 整个 Figure 的总标题
\label{fig:3B} % 整个 Figure 的标签
\vspace{-1em}
\end{figure*}

%% file: tex_figure/3B.tex
% \begin{figure*}[h] % [H] 强制放在这里, [htbp] 是更常用的选项 (here, top, bottom, page)
% \centering
% \includegraphics[width=0.95\linewidth]{figures/3B.pdf}
% \caption{Latency-aware TTS curves of s1.1-3B on MATH-500 and GPQA-Diamond.} % 整个 Figure 的总标题
% \label{fig:3B} % 整个 Figure 的标签

% \end{figure*}

%% file: tex_table/greedy_search.tex
\begin{table}[h]
\setlength{\tabcolsep}{1pt} % 将列间距从 9pt 减小到 5pt (你可以尝试 4pt, 6pt 等)
\centering
\small % 减小表格整体字体大小，可选 \footnotesize, \scriptsize
\begin{tabular}{lccc|ccc}
\toprule
\multirow{2}{*}{\textbf{Model}} & \multicolumn{3}{c|}{\textbf{Grid Search}} & \multicolumn{3}{c}{\textbf{Greedy Search}} \\
\cmidrule(lr){2-4} \cmidrule(lr){5-7}
 & $(B^*,\gamma^*)$ & Accuracy & Steps & $(B^*,\gamma^*)$ & Accuracy & Steps \\
\midrule
s1.1-32B       & $(16,5)$ & 82.3\% & 56 & $(16,5)$ & 82.3\% & 10 \\
s1.1-3B        & $(32,5)$ & 72.4\% & 56 & $(32,5)$ & 72.4\% & 11 \\
DS-32B         & $(16,4)$ & 62.2\% & 56 & $(16,4)$ & 62.2\% & 9 \\
QwQ-32B        & $(4,5)$  & 94.5\% & 56 & $(4,5)$  & 94.5\% & 8 \\
\bottomrule
\end{tabular}
\caption{Comparison of Grid Search and Greedy Search for Different Models. DS-32B: DeepSeek-Distill-32B.}
\label{tab:search_comparison}
\end{table}

%% file: tex_table/confidence_based.tex
\begin{table}[ht]
\setlength{\tabcolsep}{1pt} % 减小列间距，原为 12pt
\centering
\small % 减小表格整体字体大小，可选 \footnotesize, \scriptsize
\begin{tabular}{lcccc}
\toprule
\multirow{2}{*}{Aggregation Method} & \multicolumn{1}{c}{1024} & \multicolumn{1}{c}{2048} & \multicolumn{1}{c}{4096} & \multicolumn{1}{c}{8192} \\
\cmidrule(lr){2-2} \cmidrule(lr){3-3} \cmidrule(lr){4-4} \cmidrule(lr){5-5}
 & Acc.~(\%) & Acc.~(\%) & Acc.~(\%) & Acc.~(\%) \\
\midrule
\rowcolor{gray!20} \textbf{Majority Voting} & 82.3{\scalebox{0.6}{$\pm$1.8}} & 85.9{\scalebox{0.6}{$\pm$0.7}} & 91.7{\scalebox{0.6}{$\pm$1.3}} & 92.7{\scalebox{0.6}{$\pm$0.3}} \\
Confidence min-max & 79.2{\scalebox{0.6}{$\pm$1.6}} & 82.2{\scalebox{0.6}{$\pm$1.5}} & 88.6{\scalebox{0.6}{$\pm$1.4}} & 90.1{\scalebox{0.6}{$\pm$1.3}} \\
Confidence avg-max & 81.2{\scalebox{0.6}{$\pm$1.4}} & 83.2{\scalebox{0.6}{$\pm$1.4}} & 90.6{\scalebox{0.6}{$\pm$0.3}} & 91.1{\scalebox{0.6}{$\pm$1.2}} \\
Confidence min-vote & 82.4{\scalebox{0.6}{$\pm$1.3}} & 85.2{\scalebox{0.6}{$\pm$1.5}} & 91.5{\scalebox{0.6}{$\pm$1.6}} & 92.7{\scalebox{0.6}{$\pm$0.4}} \\
\rowcolor{gray!20} \textbf{Confidence avg-vote} & \textbf{83.0}{\scalebox{0.6}{$\pm$1.2}} & \textbf{86.3}{\scalebox{0.6}{$\pm$1.3}} & \textbf{92.3}{\scalebox{0.6}{$\pm$1.2}} & \textbf{93.5}{\scalebox{0.6}{$\pm$0.8}} \\
\bottomrule
\end{tabular}
\caption{Performance comparison of majority voting and different confidence-based strategies. Acc.: Accuracy. Results are obtained from 3 repeated experiments with mean and standard deviation reported.}
\label{tab:aggregation}
\end{table}

%% file: sec/5_conclusion.tex
\section{Conclusion}
In this paper, we propose to rethink test-time scaling in latency-sensitive scenarios. We show that compute-optimal does not always result in latency-optimal under such conditions due to memory-bound constraint. To address this, we propose to improve TTS on generation concurrency to maximize throughput from a unified view. Specifically, we present two approaches: (1) branch-wise parallelism via multiple branches and (2) sequence-wise parallelism by speculative decoding, along with their combinations. Furthermore, we investigate the concurrency allocation strategy to balance these approaches for latency-optimal TTS. A simple yet effective greedy search algorithm is proposed to determine the optimal configuration. Experimental results show that latency-optimal TTS enables 32B model to achieve 82.3\% accuracy on MATH-500 within 1 minute, while 3B model attains notable accuracy within just 10 seconds, showing significant improvements in both speed and accuracy.
% \clearpage
\section{Limitations}
\paragraph{Workload} The basic concept of this paper is that the inference of LLM is a memory-bound process. However, this concept holds on small (like mobile phones, personal computers) or medium (like work stations) scale hardware. For large-scale servers, which deals with hundreds or thousands of requests at one time, the main bottleneck of inference is computation, and the budget of inference can be measured by \#tokens. However, the studies on small and medium scale hardware still hold significant meaning, as the practical budget measured on these platforms is often under memory-bound scenarios.

\section{Potential Risks}
While our work shows test-time scaling under latency budget and our methods can significantly reduce inference latency, it introduces several risks. For instance, TTS could sometimes degrade performance on underrepresented data domains, exacerbating fairness issues. Scaled models may become more susceptible to adversarial prompts that exploit simplified decision pathways.

\section{License For Artifacts}
The artifacts utilized in this study, including datasets, codebase, and pre-trained models, are sourced from publicly available repositories under permissive licenses. All datasets adhere to open-access licenses (MIT License), ensuring compliance with redistribution and modification terms. Codebase adheres to open-access licenses (MIT License). For pre-trained models, we verify compatibility with Apache License 2.0. This alignment guarantees ethical reuse while maintaining transparency in our methodology.

\section{Information About Use of AI Assistants}
In the preparation of this work, we employ AI assistants to assist with refining academic language, and debugging code segments. The AI tools were used solely for improving clarity, grammatical correctness, and syntactic efficiency—tasks analogous to those performed by a human editor or linter. All conceptual contributions, technical claims, and critical analysis remain the authors’ own.

%% file: sec/X_appendix.tex
\clearpage
\section{Appendix}

\subsection{Why does the parallel scaling curve exhibit a steeper slope than the sequential scaling?}
This result can be attributed to the heavy model weights of LLM under the measurement of memory access. For instance, Qwen2.5-32B-Instruct has 32 billion parameters, so the model weights is 64GB. While the total KV cache at sequence length 1024 and batch size 1 is 0.25GB. Under this condition, KV cache only accounts for a small portion of memory access. When parallel scaling can easily extend branches with a small budget by 0.25GB per branch with 1024 length, sequential scaling tries hard to load the whole model weights of the entire 64GB just for 1 more extended token. This extreme imbalance on memory access makes parallel scaling significantly advantageous regarding memory access. In contrast, sequential scaling suffers from high memory access costs.

\subsection{Additional Results}
\begin{figure*}[h] % [H] 强制放在这里, [htbp] 是更常用的选项 (here, top, bottom, page)
\centering
\includegraphics[width=\linewidth]{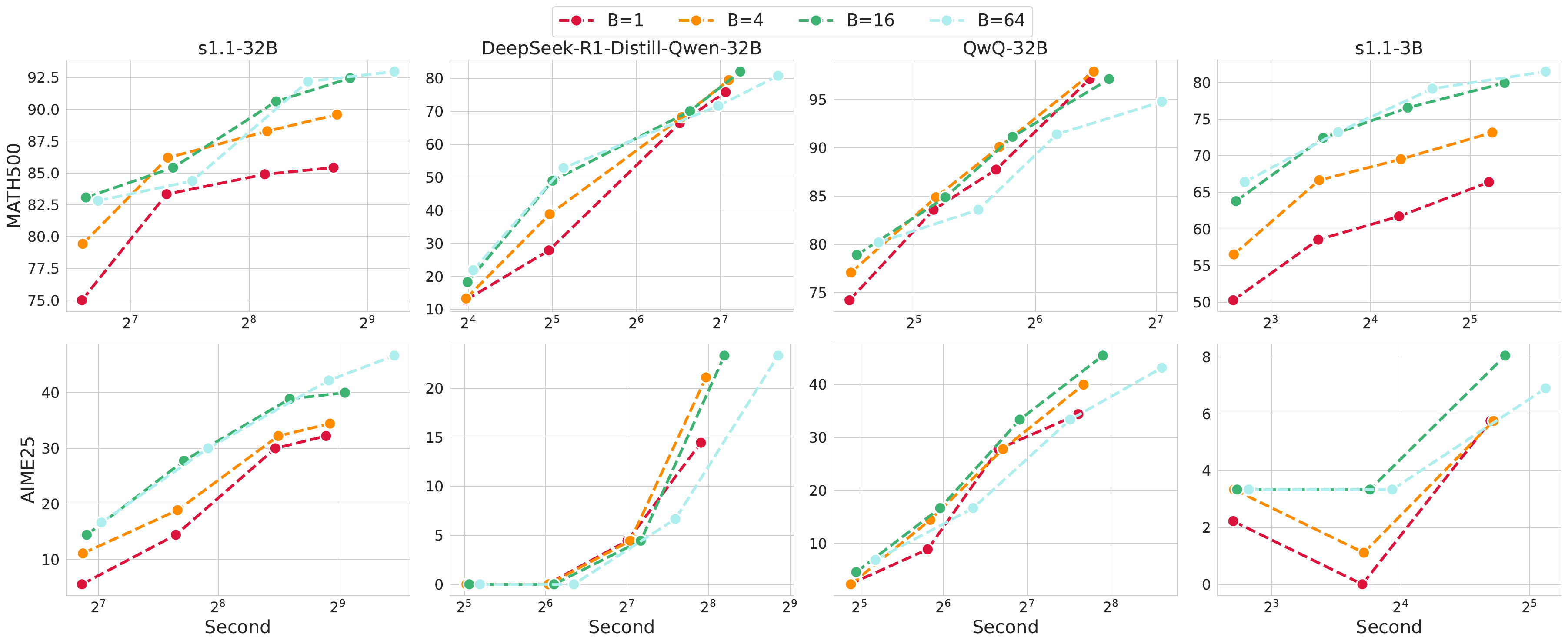}
\caption{More results of the influence of the number of branches from branch-wise parallelism.} % 整个 Figure 的总标题
\label{fig:branch_supp} % 整个 Figure 的标签

\end{figure*}

\begin{figure*}[h] % [H] 强制放在这里, [htbp] 是更常用的选项 (here, top, bottom, page)
\centering
\includegraphics[width=\linewidth]{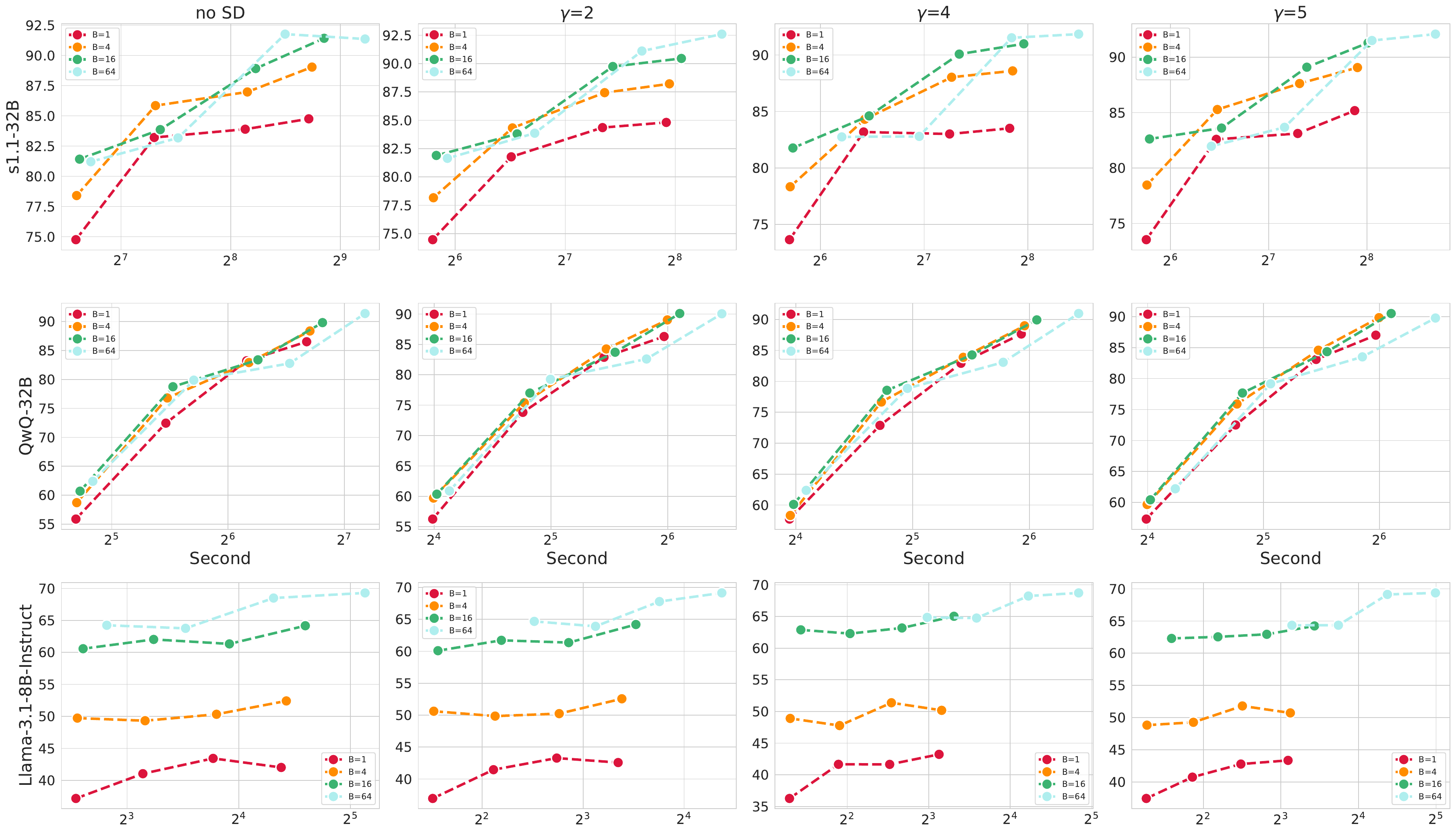}
\caption{More results of the influence of the number of branches from branch-wise parallelism under different sequence-wise configurations. The draft lengths of speculative decoding varies among \{2, 4, 5\}.} % 整个 Figure 的总标题
\label{fig:trade_off_branch_supp} % 整个 Figure 的标签

\end{figure*}

\begin{figure*}[h] % [H] 强制放在这里, [htbp] 是更常用的选项 (here, top, bottom, page)
\centering
\includegraphics[width=\linewidth]{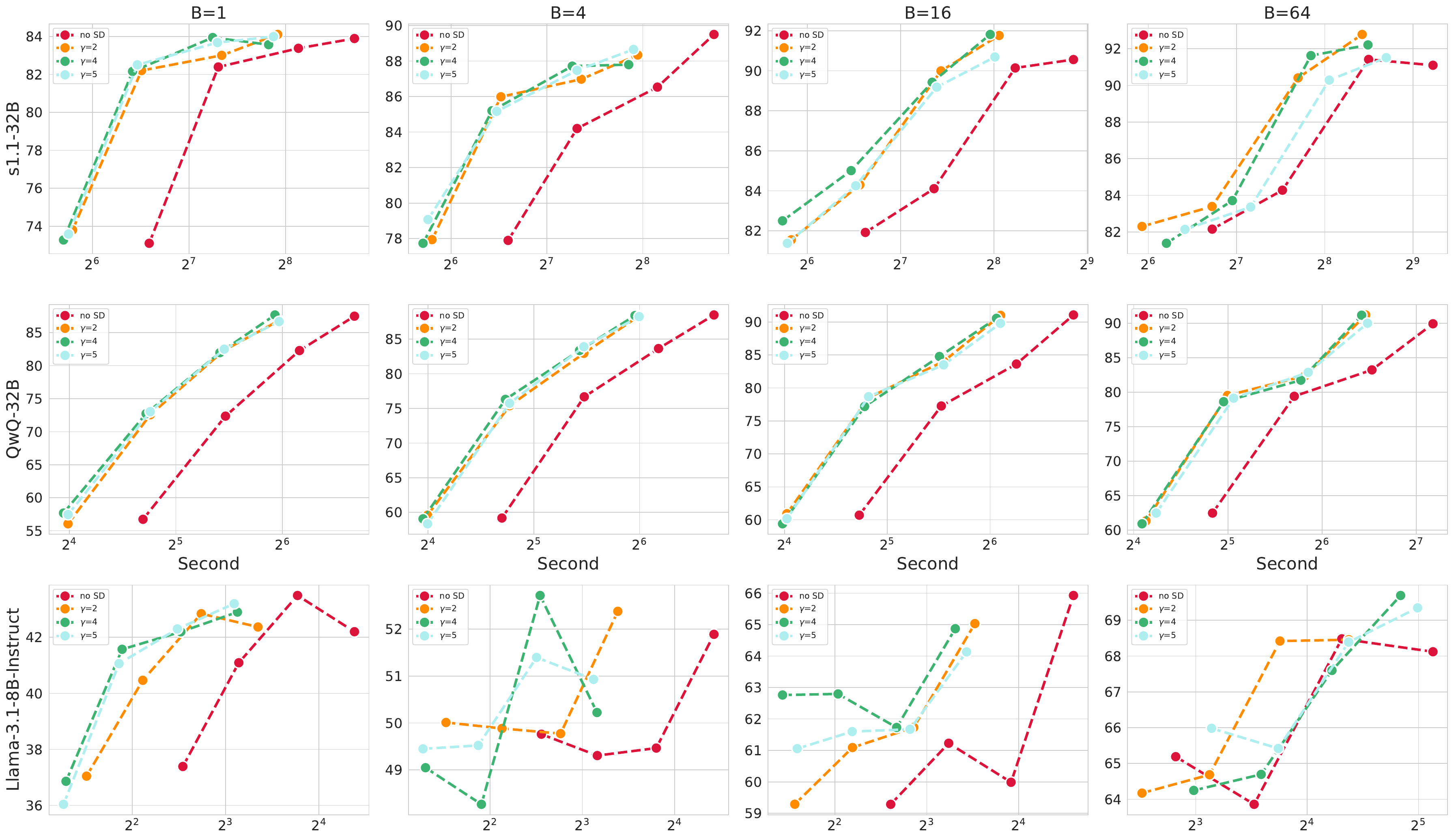}
\caption{More results of the influence of draft length from sequence-wise parallelism under different branch-wise configurations. The number of branches varies among \{1, 4, 16, 64\}.} % 整个 Figure 的总标题
\label{fig:trade_off_sd_supp} % 整个 Figure 的标签
\end{figure*}

We present additional experimental results. The results on the influence of branch-wise parallelism is shown in Figure~\ref{fig:branch_supp}. The results of the influence of branch-wise parallelism under different speculative decoding configurations is shown in Figure~\ref{fig:trade_off_branch_supp}. The results of sequence-wise parallelism under different branch counts are shown in Figure~\ref{fig:trade_off_sd_supp}..

The acceptance rate $\alpha$ of the speculative decoding we employ is reported below: Target: s1.1-32B, draft: s1.1-7B, $\alpha$: 0.831. Target: DeepSeek-R1-Distill-Qwen-32B, draft: DeepSeek-R1-Distill-Qwen-7B, $\alpha$: 0.897. Target: QwQ-32B, draft: DeepSeek-R1-Distill-Qwen-7B, $\alpha$: 0.781. Target: LLaMa-3.1-8B-Instruct, draft: Eagle3, $\alpha$: 0.904. Target: s1.1-3B, draft: Qwen2.5-0.5B-Instruct, $\alpha$: 0.701. 

The latency-optimal configurations shown in Figure~\ref{fig:latencyoptimal1}, ~\ref{fig:latencyoptimal2}, ~\ref{fig:32B} and~\ref{fig:3B} are reported below: Figure~\ref{fig:latencyoptimal1}: Left: B=16, $\gamma$=5. Right: B=32, $\gamma$=3. Right: B=32, $\gamma$=5. Figure~\ref{fig:latencyoptimal2}: AIME24 with QwQ-32B: B=16, $\gamma$=4. AIME24 with DeepSeek-R1-Distill-Qwen-32B: B=32, $\gamma$=4. AIME25 with QwQ-32B: B=16, $\gamma$=4. AIME24 with DeepSeek-R1-Distill-Qwen-32B: B=32, $\gamma$=5. Figure~\ref{fig:32B}: MATH-500: B=16, $\gamma$=5. GPQA-Diamond: B=32, $\gamma$=5. AIME24: B=32, $\gamma$=4. AIME25: B=32, $\gamma$=5. Figure~\ref{fig:3B}: MATH-500: B=32, $\gamma$=5. GPQA-Diamond: B=16, $\gamma$=5. AIME24: B=8, $\gamma$=5. AIME25: B=16, $\gamma$=5.

\input{tex_table/multi_request}

% \begin{figure*}[h] % [H] 强制放在这里, [htbp] 是更常用的选项 (here, top, bottom, page)
% \centering
% \includegraphics[width=\linewidth]{figures/8_tradeoff.pdf}
% \caption{Results of different branch-wise parallelism and sequence-wise parallelism configurations with fixed generation concurrency.} % 整个 Figure 的总标题
% \label{fig:8_tradeoff} % 整个 Figure 的标签
% \end{figure*}

\subsection{Discussion on prefill phase}
The whole inference time of LLM is composed of prefilling and decoding. While prefill time is indeed an important overhead of the overall inference, it is not the dominant factor in our case. For example, we have analyzed MATH-500 and find that the question length distribution is highly skewed towards shorter inputs: 88\% of questions contain fewer than 360 tokens, and 99\% contain fewer than 800 tokens. In contrast, our model generates responses up to 8,192 tokens in length. On modern GPUs, the prefill phase is significantly faster than the autoregressive decoding phase. Given the ratio of input and output lengths in our dataset, the prefill time is negligible of the total latency.

Therefore, while prefill time is technically part of the inference process, its contribution is small. Excluding it does not affect the conclusions.

\subsection{Discussion on large batch, multi-requests scenario}
We show the experiments as number of branches grows in the table below. The first column: model configuration. The second column: Accuracy. \#Req: number of requests. Each cell contains the latency. With more requests, the workload transitions to a compute-bound state. At this point, latency depends on the FLOPs, which is directly proportional to the token count. Consequently, for our latency-aware strategy, once the workload transitions from memory-bound to compute-bound, the trade-off strategies can be directly informed by previous token count strategy.

%% file: tex_table/multi_request.tex
\begin{table*}[h]
\centering
\begin{tabular}{lcc ccc}
\toprule
\multirow{2}{*}{Models} & \multirow{2}{*}{Configuration} & \multirow{2}{*}{Accuracy} & \multicolumn{3}{c}{\textbf{\#Req}} \\ 
\cmidrule(lr){4-6}
& & & \textbf{1} & \textbf{4} & \textbf{16} \\
\midrule
s1.1-32B & $B=1, \gamma=5$ & 82.3\% & \cellcolor{gray!20}90.5s & \cellcolor{gray!20}91.2s & \cellcolor{gray!20}92.9s \\
s1.1-32B & $B=4, \gamma=5$ & 85.6\% & \cellcolor{gray!20}91.3s & \cellcolor{gray!20}93.1s & 156.8s \\
s1.1-32B & $B=16, \gamma=5$ & 83.8\% & \cellcolor{gray!20}90.5s & 142.7s & 576.8s \\
s1.1-32B & $B=64, \gamma=5$ & 83.0\% & 159.8s & 602.8s & 2621.8s \\
Llama-3.1-8B-Instruct & $B=1, \gamma=4$ & 42.3\% & \cellcolor{gray!20}7.6s & \cellcolor{gray!20}7.6s & \cellcolor{gray!20}8.2s \\
Llama-3.1-8B-Instruct & $B=4, \gamma=4$ & 47.6\% & \cellcolor{gray!20}7.6s & \cellcolor{gray!20}8.3s & 26.4s \\
Llama-3.1-8B-Instruct & $B=16, \gamma=4$ & 63.1\% & \cellcolor{gray!20}8.1s & 25.8s & 95.8s \\
Llama-3.1-8B-Instruct & $B=64, \gamma=4$ & 64.9\% & 24.4s & 91.2s & 374.5s \\
\bottomrule
\end{tabular}
\caption{Latency under different configurations and number of requests (\#Req). Memory-bound workloads are marked with gray. As the number of computed tokens per forward grows, the system transits from memory-bound to compute-bound. The latency of the latter can be measured by \#tokens.}
\label{tab:multi_request}
\end{table*}